%% file: main.tex
\documentclass[11pt,a4paper,x11names]{article}
\usepackage{setspace}
\usepackage{misc/acl}
\usepackage{times}
\usepackage[table]{xcolor}
\usepackage{graphicx}
\usepackage{xfrac} %
\usepackage[T1]{fontenc}
\usepackage[utf8]{inputenc}
\usepackage{microtype}
\usepackage{booktabs}
\usepackage{todonotes}
\usepackage{amsmath,amssymb,bbm}
\usepackage{cleveref}
\usepackage{xspace}
\usepackage{listings,algorithm, algorithmicx, algpseudocode}
\usepackage{soul} %
\usepackage[hypcap=false]{caption}
\usepackage{multirow,multicol}
\usepackage{ulem}
\usepackage{float}
\usepackage{array}
\usepackage{bm}
\usepackage{fancyvrb} %
\usepackage{xfrac}
\usepackage[shortlabels]{enumitem}
\usepackage{dsfont} %
\usepackage{url} %
\usepackage{tcolorbox}
\usepackage{makecell} %
\DeclareUrlCommand\UScore{\urlstyle{rm}}
\usepackage{anyfontsize} %
\usepackage{siunitx}    %
\usepackage{adjustbox}  %
\usepackage{pifont}
\usepackage{siunitx}
\usepackage{rotating} %

\usepackage{silence}
\usepackage{tabularx}

\usepackage{stmaryrd}
\usepackage{trimclip}

\crefname{lstlisting}{listing}{listings}
\Crefname{lstlisting}{Listing}{Listings}
\crefname{myequation}{equations}{equations}
\Crefname{myequation}{Equations}{Equations}
\Crefname{algorithmCaption}{Algorithm}{Algorithms}
\crefname{example}{example}{examples}
\Crefname{example}{Example}{Examples}
\crefname{prompt}{prompt}{prompts}
\Crefname{prompt}{Prompt}{Prompts}
\DeclareCaptionType{algorithmCaption}[Algorithm][List of algorithms]
\DeclareCaptionType{example}[Example][List of examples]
\DeclareCaptionType{prompt}[Prompt][List of prompts]
\DeclareCaptionType{myequation}[Equation][List of equations]

\usepackage{courier}
\usepackage{marginnote}

\input{misc/macros.tex}

\definecolor{easygreen}{HTML}{DFF0D8}
\definecolor{hardred}{HTML}{F8D7DA}
\definecolor{mixedorange}{HTML}{FFE5B4}

\newcommand{\langpair}[2]{\textsc{#1}$\rightarrow$\textsc{#2}}

\usepackage{tcolorbox}
\tcbuselibrary{listingsutf8}

\author{
Lorenzo Proietti$^{1,*}$
\quad Stefano Perrella$^{1,*}$
\quad  Vilém Zouhar$^{2,*}$ \\
\bf Roberto Navigli$^1$ \quad Tom Kocmi$^3$ \\
$^1$Sapienza NLP Group, Sapienza University of Rome
\quad $^2$ETH Zurich
\quad $^3$ Cohere \\
\small\tt
\{\href{mailto:lproietti@diag.uniroma1.it}{\color{black}lproietti},\href{mailto:perrella@diag.uniroma1.it}{\color{black}perrella},\href{mailto:navigli@diag.uniroma1.it}{\color{black}navigli}\}@diag.uniroma1.it \quad
\href{mailto:vzouhar@ethz.ch}{\color{black} vzouhar@ethz.ch} \quad
\href{mailto:kocmi@cohere.com}{\color{black} kocmi@cohere.com}
}

\title{Estimating Machine Translation Difficulty}

\begin{document}

\maketitle

\def\thefootnote{*}\footnotetext{Equal contribution.}
\def\thefootnote{\arabic{footnote}}

\input{content.tex}

\bibliography{misc/bibliography.bib,misc/anthology.min.bib}
\bibliographystyle{misc/acl_natbib}

\clearpage

\appendix
\input{appendix.tex}

\end{document}

%% file: misc/macros.tex
\definecolor{TodoColor}{rgb}{1,0.7,0.6}

\newcommand{\sentinel}{$\text{Sentinel-src}$\xspace}
\newcommand{\sentineltf}{$\text{Sentinel-src-24}$\xspace}
\newcommand{\sentinelwmt}{$\text{Sentinel-src-25}$\xspace}

\newcommand{\systems}{\ensuremath{\mathcal{M}_l}\xspace}
\newcommand{\texts}{\ensuremath{\mathcal{X}}\xspace}

\definecolor{FindingsColor}{gray}{0.85}

\definecolor{darkred}{RGB}{139,0,0}
\definecolor{darkgreen}{RGB}{0,100,0}
\newcommand{\tick}{\includegraphics[width=0.65em]{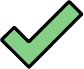}}
\newcommand{\cross}{\includegraphics[width=0.65em]{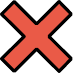}}

\makeatletter\def\Hy@Warning#1{}\makeatother
\let\svthefootnote\thefootnote
\newcommand\blankfootnote[1]{%
  \let\thefootnote\relax\footnotetext{#1}%
  \let\thefootnote\svthefootnote%
}

%% file: content.tex
\begin{abstract}
Machine translation quality has steadily improved over the years, achieving near-perfect translations in recent benchmarks.
These high-quality outputs make it difficult to distinguish between state-of-the-art models and to identify areas for future improvement.
In this context, automatically identifying texts where machine translation systems struggle holds promise for developing more discriminative evaluations and guiding future research.

In this work, we address this gap by formalizing the task of translation difficulty estimation, defining a text's difficulty based on the expected quality of its translations.
We introduce a new metric to evaluate difficulty estimators and use it to assess both baselines and novel approaches.
Finally, we demonstrate the practical utility of difficulty estimators by using them to construct more challenging benchmarks for machine translation. 
Our results show that dedicated models outperform both heuristic-based methods and LLM-as-a-judge approaches, with \sentinel achieving the best performance.
Thus, we release two improved models for difficulty estimation, \sentineltf and \sentinelwmt, which can be used to scan large collections of texts and select those most likely to challenge contemporary machine translation systems.

\end{abstract}

\smallskip
\section{Introduction}

Not all data samples are equal: Machine learning models may struggle with some samples more than others.
The ability to automatically assess sample difficulty is indispensable at various stages of model development.
For example, during training, organizing samples from the easiest to the hardest, known as Curriculum Learning, improves both performance and training efficiency \cite{cl, surveycl1, surveycl2}.
Even during inference, computational costs can be reduced by early-exiting on easy examples \citep{branchynet,schwartz-etal-2020-right}.

Evaluating models also benefits from estimates of sample difficulty, as too easy or too difficult benchmarks fail to effectively differentiate between models \cite{lalor-etal-2018-understanding,rodriguez-etal-2021-evaluation}.
This issue is particularly relevant in Machine Translation (MT), with recent state-of-the-art models obtaining near-perfect scores and performing close to the human level \cite{kocmi-etal-2024-findings, proietti-etal-2025-machine}.
With easy test sets, practitioners might struggle to differentiate between top-performing models and assess whether there is headroom for further model improvement.
Additionally, while the MT Test Suites subtask of WMT \citep{kocmi-etal-2024-findings} targets specific complex translation phenomena, no systematic investigation of the broader concept of general translation difficulty has been carried out.

To address this gap, we explore the notion of sample difficulty in machine translation.
First, we propose a definition of translation difficulty and formally introduce \textit{translation difficulty estimation} as a novel task, where the source text's difficulty is automatically predicted.
We then present \textit{difficulty estimation correlation} (DEC), a measure designed to evaluate the performance of difficulty estimation methods.
Finally, we test baselines and newly proposed approaches to difficulty estimation and validate their practical utility in the downstream application of creating a challenging benchmark, which involves automatically selecting subsets of challenging samples from a large corpus.

We find that approaches such as word rarity, syntactic complexity, or even LLM-as-a-Judge underperform dedicated solutions in capturing translation difficulty. Specifically, we find \sentinel -- a model trained to predict the expected translation quality of a given text based solely on the source text itself \cite{perrella-etal-2024-guardians} -- to outperform other methods at estimating translation difficulty.
Therefore, we train two improved versions, called \sentineltf and \sentinelwmt, and release them publicly.\footnote{Models: \href{https://huggingface.co/collections/Prosho/translation-difficulty-estimators-6816665c008e1d22426eb6c4}{hf.co/collections/Prosho/translation-difficulty-estimators-6816665c008e1d22426eb6c4}.\\
Code: \href{https://github.com/zouharvi/translation-difficulty-estimation}{github.com/zouharvi/translation-difficulty-estimation}.}

\section{Related Work}
\label{sec:related_work}

Previous works can be divided into two categories depending on whether their focus is human or machine translation difficulty. 

\paragraph{Human translation difficulty.}
The earliest works \citep{fang1959some,hale2002interaction} attempted to connect general text complexity to translation difficulty for humans.
A more modern investigation by \citet{mishra-etal-2013-automatically} framed human translation difficulty as the time needed to translate a sentence, and estimated it using data on translators' eye movements.
They use text length, word polysemy degree, and syntactic complexity as predictors of translation difficulty.
\citet{correlating_process} examined the correlation between error count, word translation entropy, and syntactic equivalence with translation duration, gaze, and other proxies for human translation difficulty.
More recently, \citet{lim-etal-2023-predicting,lim-etal-2024-predicting} used word alignment distributions and decoder perplexity to predict human translation difficulty.

\paragraph{Machine translation difficulty.}
To implement a Curriculum Learning training schedule, \citet{kocmi-bojar-2017-curriculum} estimated sample difficulty based on sentence length, word rarity, and the number of coordinating conjunctions in the text.
Similarly, \citet{platanios-etal-2019-competence} used sentence length and rarity as proxies for difficulty.
Beyond these linguistically-motivated criteria, \citet{zhang2018empiricalexplorationcurriculumlearning} and \citet{liu-etal-2020-norm} predicted translation difficulty using the confidence and other intrinsics of the translation model in generating the text.
\citet{Almeida2017DifficultyEO} treated difficulty estimation as a binary classification task, although they also used features from the target text, making it closer to quality estimation.
\citet{zhan2021varianceaware} used an artificial crowd-based approach that leverages automatic metrics and discovered that long segments, low-frequency words, and proper nouns are the most challenging to machine translate.
Finally, \citet{zhan-etal-2021-difficulty} estimated text's difficulty using the embedding similarity between its tokens and those of its translations.

Closer to our work, \citet{don-yehiya-etal-2022-prequel} defined the PreQuEL task as predicting the quality of a given text's translation before the translation is generated.
However, they adopted the evaluation of the WMT 2020 Quality Estimation Shared Task \cite{specia-etal-2020-findings}, which was designed for quality estimation rather than for assessing difficulty estimators.
Furthermore, their test set included only two language directions, with all translations produced by the same MT model.
Additionally, they did not explore the broader space of difficulty estimators or investigate their use in constructing challenging benchmarks.

In contrast, we define translation difficulty estimation as a distinct task with a dedicated evaluation metric.
Moreover, we benchmark a wide range of difficulty estimation approaches using test sets that span $11$ language directions, with $11$ to $19$ translations per segment across language pairs, produced by both MT models and human translators.
As a result, our work constitutes the first extensive evaluation of translation difficulty estimators, establishing a new state of the art for the task. 

\section{The Difficulty Estimation Task}
\label{sec:difficulty}

The difficulty of translating a given text can depend on multiple factors.
A text may be challenging, for example, due to its length, syntactic complexity, idiomatic language, or the presence of rare or specialized vocabulary.
Some aspects that affect translation difficulty may even depend on the translation direction, meaning that the same source text might be more difficult to translate into one language than into another.
Moreover, translation difficulty might not be uniform across translators, as it can vary with the translator's cultural background and linguistic familiarity -- in the case of human translators -- or based on factors such as the number of parameters, training data, and model architecture -- in the case of machine translation models.

Given these considerations, we avoid defining translation difficulty in absolute terms, as such a definition may not generalize well. Instead, we define difficulty relative to a given target language and to the accuracy of a particular translator, whether human or automatic. More specifically, given a text $x$, a model\footnote{For brevity, we use ``model'' to refer both to human translators and automatic models.} $m$, and a target language $l$, we assign to $x$ a difficulty score $d_{m, l}(x)$ equal to the quality score assigned to a translation of $x$ into language $l$ produced by $m$. Lower scores indicate a lower translation quality and, therefore, greater difficulty associated with the source text.

As an example, suppose we have two texts, $x_1$ and $x_2$, and their respective translations $t_1$ and $t_2$ into language $l$, both produced by model $m$. A human rater evaluates these translations on a scale from $1$ to $100$, assigning a score of $60$ to the first and $90$ to the second. Then, $d_{m,l}(x_1) = 60$ and $d_{m,l}(x_2) = 90$. Since $d_{m,l}(x_1) < d_{m,l}(x_2)$, then $x_1$ is more difficult to translate into $l$ than $x_2$, for model $m$.
Importantly, the lower the score $d$, the higher the difficulty and vice versa.

\paragraph{Task Definition.}
Given a source text $x$, a model $m$, and a target language $l$, Difficulty Estimation is the task of automatically predicting $d_{m, l}(x)$.
Different from Quality Estimation, difficulty estimation models do not have access to the translations whose quality is being estimated. Indeed, difficulty estimation can be seen as the task of estimating the \textit{expected} quality of a given text's translation.

\paragraph{Evaluation.}
We evaluate difficulty estimation methods according to their ability to rank texts based on difficulty.
Consider a collection of texts $\texts=x_1, x_2, \dots,$ $x_N$, a collection of target languages $\mathcal{L} = l_1,$ $l_2,\dots,l_L$, and a collection of models translating into language $l$: $\mathcal{M}_l=m_1, m_2, \dots, m_{M_l}$.
Let us also define the vector of ground-truth difficulty scores for model $m$ and language $l$ as $D_{m,l}=d_{m,l}(x_1),$ $d_{m,l}(x_2), \dots, d_{m,l}(x_N)$ and the corresponding predictions of a difficulty estimation method as $\hat{D}_{m,l}=\hat{d}_{m,l}(x_1),$ $\hat{d}_{m,l}(x_2),$ $\dots,$ $\hat{d}_{m,l}(x_n)$.
We measure the translation \textbf{Difficulty Estimation Correlation (DEC)} by averaging the Kendall's rank correlation coefficients $\tau_b$ across models and languages:
\begin{equation}
    \text{DEC} \;=\; \frac{1}{|\mathcal{L}|} \sum_{l\in\mathcal{L}} \frac{1}{|\systems|} \sum_{m \in \systems}\tau_b\!\ \bigl(\hat{D}_{m,l}, D_{m,l}\bigr). \label{eq:dec}
\end{equation}
We refer the reader to \Cref{sec:kendall} for further details on how the Kendall correlation coefficient $\tau_b$ is computed.

\paragraph{Contrasting DEC with standard MT meta-evaluation strategies.}
The evaluation approach used in \ref{eq:dec}, termed ``Group-by-System'' by \citet{deutsch-etal-2023-ties}, makes DEC fundamentally different from other meta-evaluation strategies used in MT evaluation and Quality Estimation, which typically rely on the Group-by-Item method instead \cite{deutsch-etal-2023-ties}. Group-by-Item calculates the correlation between assessments assigned to different translations of the \textbf{same source text} (i.e., the same evaluation item), and then averages these correlations across all source texts. The primary benefit of this method is that it mitigates spurious correlations between source text features (e.g., length) and translation quality judgments \cite{perrella-etal-2024-guardians}. 

However, we argue that these features are precisely what define a text's translation difficulty. Since our goal is to measure translation difficulty, we define DEC using Group-by-System. This method computes the correlation between human and metric assessments for translations of \textbf{different source texts} that were produced by the same MT system, and then averages the correlations across MT systems. By holding the MT system constant, this evaluation directly measures a metric's ability to identify which source texts were more challenging for that system to translate.

This distinction is crucial. Indeed, using Group-by-System to meta-evaluate standard MT or quality estimation metrics would favor those that predict source difficulty rather than purely translation quality. Conversely, using Group-by-Item to evaluate difficulty estimators would be inappropriate, as these estimators assign the same score to all translations of a given source.

\section{Methods for Difficulty Estimation} \label{sec:methods}

In this section, we describe several difficulty estimation methods.
We include both common and novel approaches.
Specifically, we categorize difficulty estimators into four groups: heuristic-based, learned, LLM-as-a-Judge, and artificial crowd-based.
We refer the reader to \Cref{apx:implementation_details} for implementation details regarding all the considered models.

\subsection{Heuristic-based estimators}
We refer to estimators as heuristic-based if they rely on simple text features.
This category includes estimators previously shown to correlate with other measures of difficulty \cite{mishra-etal-2013-automatically, kocmi-bojar-2017-curriculum, Araghi2024}.
\begin{itemize}
    \item \textbf{Text length} is the number of tokens in a text.

    \item \textbf{Word rarity} is the negative average of the frequencies (estimated from a reference corpus) of the words in a text.

    \item \textbf{Syntactic complexity} is approximated as the height of the dependency tree associated with a text.
\end{itemize}

\subsection{Learned estimators}

Learned machine translation metrics are often trained to predict the quality of a translation given its source text and, optionally, a reference translation \cite{rei-etal-2020-comet,guerreiro2024xcomet,juraska-etal-2023-metricx,juraska-etal-2024-metricx}.
Similarly, neural models can be trained to predict the difficulty of a text.
Previous research has explored training similar models for related purposes: 

\begin{itemize}

    \item \textbf{PreCOMET} is a suite of source-based regressors based on XLM-RoBERTa \cite{conneau-etal-2020-unsupervised} that predict the usefulness of a sample for evaluation \citep{zouhar2025subset}. Specifically, PreCOMET\textsubscript{diversity} prioritizes samples likely to elicit diverse machine translation outputs, while
    PreCOMET\textsubscript{difficulty} estimates difficulty as defined by item response theory \citep{santor1998progress}.

    \item \textbf{\sentinel} metric is a regression model based on XLM-RoBERTa. \citet{perrella-etal-2024-guardians} trained \sentinel to estimate translation quality from the source alone -- i.e., without accessing the candidate translation -- with the goal of learning spurious correlations between features of the source texts and translation quality scores.
\end{itemize}

\subsection{LLM-as-a-Judge} \label{sec:llm-as-a-judge}
LLM-as-a-Judge approaches have seen wide adoption across a range of applications \cite{zheng2023judgingllmasajudgemtbenchchatbot, bavaresco2024llmsinsteadhumanjudges}. In this work, we investigate the effectiveness of the LLM-as-a-Judge paradigm for the task of difficulty estimation, using GPT-4o \cite{openai2024gpt4technicalreport} and CommandA \cite{cohere2025commandaenterprisereadylarge}.
We prompt these models to determine the proficiency level required to translate a given text, optionally providing information about the target language, and return a scalar score between 0 and 120 indicating the difficulty level of the given text.
See the prompts in \Cref{fig:llm_prompts}.

\subsection{Crowd-based Estimators} \label{sec:crowd-based-estimators}
The methods discussed so far estimate translation difficulty based solely on the source text, and optionally, the target language. However, having defined translation difficulty as the expected quality of a model's translations (Section~\ref{sec:difficulty}), we now introduce difficulty estimators that more closely mimic this definition. 

\paragraph{Artificial Crowd.}
Artificial crowd-based methods first translate a source text and then use reference-less MT metrics to estimate the quality of the resulting translations.\footnote{Reference-less MT metrics estimate the quality of a translation by comparing it only to its source text, without requiring reference translations.} Specifically, we translate the source texts from the test set using a diverse set of models to ensure variety in architecture and size: three instruction-tuned LLMs (Gemma-3-27B-IT, Qwen2.5-72B-IT, CommandA) and one standard encoder-decoder machine translation model (NLLB-moe-54B). For the evaluation step, we employ two state-of-the-art, reference-less MT metrics: XCOMET-QE-XXL \cite{guerreiro-etal-2024-xcomet} and MetricX-24-Hybrid-QE-XXL \cite{juraska-etal-2024-metricx}, hereafter referred to as XCOMET and MetricX, respectively. The final difficulty score for each source text is the average quality score assigned to its translations by one of these metrics. This approach is inspired by the artificial crowd methods for efficient subset selection proposed by \citet{zouhar2025subset}. 

\paragraph{True Crowd.}
To establish a performance upper bound for Artificial Crowd estimators, we also define True Crowd estimators. Unlike Artificial Crowd, True Crowd estimators use XCOMET and Metric to score the translations produced by the actual systems whose difficulty we aim to measure -- namely, the translations that constitute the WMT24 test set used in our experiments. 

Since they rely on the ``ground-truth'' translations, True Crowd estimators are effectively equivalent to quality estimators. Thus, they are not proper difficulty estimators; we employ them solely to report an upper bound on the performance of Artificial Crowd estimators.

\section{Experiments}
\label{sec:experiments}

We benchmark the estimators using the difficulty estimation correlation measure (DEC, Equation~\ref{eq:dec}).

\subsection{Experimental Setup}
We measure DEC on the test sets released at the WMT 2024 General MT and Metrics shared tasks \cite{kocmi-etal-2024-findings, freitag-etal-2024-llms}. These test sets include a selection of source texts translated into multiple languages by automatic models and human translators. Each translation is paired with quality annotations produced by human annotators following either the Error Span Annotation (ESA, \citealp{kocmi-etal-2024-error}) or the Multidimensional Quality Metrics (MQM, \citealp{lommel2014multidimensional, freitag-etal-2021-experts}) annotation protocols.
Here we report results with the ESA annotation protocol.
See Appendix \Cref{tab:dec_results_mqm} for results with the MQM protocol, and Appendix \Cref{tab:wmt24_stats_esa,tab:wmt24_stats_mqm} for data statistics.

We test all methods listed in \Cref{sec:methods}.
Additionally, we improve the top-performing learned estimator, \sentinel, by expanding the training data used by \citet{perrella-etal-2024-guardians} and training two new models, dubbed \sentineltf and \sentinelwmt.
The former is trained with data from previous WMT editions up to WMT 2023, while the latter also includes the WMT 24 test set.\footnote{For this reason, \sentinelwmt is not included in the results in Table~\ref{tab:dec_results_esa}.}
See \Cref{apx:sentinel-training} for further details regarding the training pipeline and parameters of \sentineltf and \sentinelwmt.

\input{tables/dec_results_esa}

\subsection{Results} \label{sec:results}

We present the results in \Cref{tab:dec_results_esa}, with methods organized by category as described in \Cref{sec:methods}.
We also mark each method with the information it uses (i.e., true translations or target language), as detailed in the caption of \Cref{tab:dec_results_esa}, and include three distinct oracles to provide the reader with upper-bound performance values. 
The definition of Oracles can be found in \Cref{apx:oracles}.

\paragraph{Heuristic-based and Learned methods.}
These estimators base their predictions only on the input text.
Consequently, the difficulty scores they assign to each text are the same across all target languages and models.\footnote{I.e., $\forall m_1,m_2{\in}\mathcal{M},  l_1,l_2{\in}\mathcal{L}: \, \hat{d}_{m_1,l_1}(x) = \hat{d}_{m_2,l_2}(x)$.}
Within this group, all learned estimators outperform the heuristic-based ones. Furthermore, \sentineltf achieves the highest difficulty estimation correlation overall, also higher than \sentinel from \citet{perrella-etal-2024-guardians}, highlighting the effectiveness of our re-training.

\paragraph{LLM-as-a-Judge.}
LLM judges are optionally provided with the target language.
For both models, the target language information improves performance. This is especially true for CommandA, where the target language information leads to a $0.032$ points increase in correlation. However, the overall LLM judges' performance is poor, with scores even lower than the much simpler Text Length heuristics.  

\paragraph{Crowd-based estimators.}
As expected, the True Crowd methods, which utilize ground-truth translations and thus serve as an upper bound, yield the highest correlation. Since their performance depends solely on the employed reference-less metrics, this result also demonstrates that XCOMET outperforms MetricX on this task by a noticeable margin. 

Instead, Artificial Crowd methods' performance is comparable to that of \sentineltf.
However, Artificial Crowd approaches are considerably more resource-intensive than learned methods, as they require both the translation of source texts and a subsequent quality estimation step.

\subsection{Discussion} 
Through our evaluation, we find that:
\begin{itemize}
    \item Heuristic-based estimators, commonly used in previous works, are outperformed by most other methods.

    \item LLM-as-a-Judge approaches are also surpassed by most methods, including the much simpler Text Length heuristic.

    \item The performance of learned methods -- i.e., models explicitly trained to predict text difficulty -- is matched only by Artificial Crowd estimators, which are, however, considerably more expensive to operate. 

    \item \textbf{\sentineltf sets a state-of-the-art in difficulty estimation}, outperforming all other evaluated estimators.\footnote{We intentionally exclude True Crowd estimators, which, as discussed in \Cref{sec:crowd-based-estimators}, are not genuine difficulty estimators and instead serve only as upper bounds for Artificial Crowd.}
\end{itemize}

Based on these results, \Cref{sec:difficult_subset} examines the ability of the top estimators from each category -- excluding True Crowd estimators, for the reasons discussed in \Cref{sec:crowd-based-estimators} -- on the downstream task of constructing difficult benchmarks. 

Furthermore, even if Artificial Crowd estimators are included, using them for benchmark creation implicitly assumes that the generated test data will be used to evaluate models other than those involved in the difficulty estimation. In fact, because Artificial Crowd relies on an intermediate translation step, it would bias the resulting test set against the models used in its construction.

Full results with significance testing -- including those restricted to the MQM-annotated portion of WMT24 -- are provided in \Cref{apx:complete_results}.

\input{tables/human_mt_difficulty}

\subsection{Comparing Human and Machine Translation Difficulty}

In this section, we examine whether the texts that models find difficult to translate are also challenging for humans. 
To do this, we use the difficulty scores $d_{m,l}$ assigned to each source text in the WMT 24 test sets, varying $m$ across human translators and MT models.
We measure the Kendall's $\tau_b$ between the scores of all pairs of translators, averaging them across all language directions.
For consistency, we restrict the analysis to the models and human translators for which we have annotated translations for all language directions, namely, one human translator and the following four models: Unbabel-Tower70B \cite{alves2024tower}, IOL-Research \cite{zhang-2024-iol}, Claude-3.5, and GPT-4 \cite{openai2024gpt4technicalreport}.

The results in \Cref{tab:avg_corr_matrix_human_mt} show that the correlations with the human translator are consistently lower (ranging $0.109$ to $0.137$) than those between machine translation models (ranging $0.151$ to $0.221$).
This suggests that human translators may perceive translation difficulty differently from automatic models.
Notably, the highest agreement is observed between GPT-4 and Claude-3.5, which might be due to both models being general-purpose LLMs, unlike Unbabel-Tower70B and IOL-Research, which were explicitly trained for machine translation.

\begin{figure}[t]
\centering
    \includegraphics[width=\linewidth]{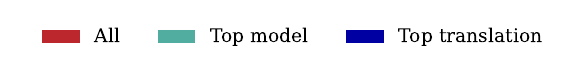}

    \vspace{-2mm}
    \includegraphics[width=\linewidth]{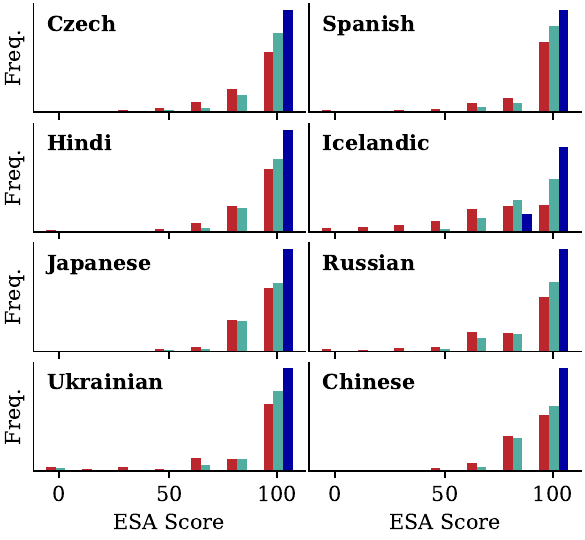}

    \vspace{-5pt}

    \caption{Distribution of human scores assigned to the translations of the texts included in the WMT 2024 test set \citep{kocmi-etal-2024-findings}. We report the scores of all models (ALL), the scores of the top-performing model for each language (Top model), and the scores of the best translation for each input text (Top translation). The chosen bin width is 15 ESA points.}
    \label{fig:07-perlang_wmt24}
\end{figure}

\begin{figure*}[t]
    \centering
    
    \includegraphics[width=\linewidth]{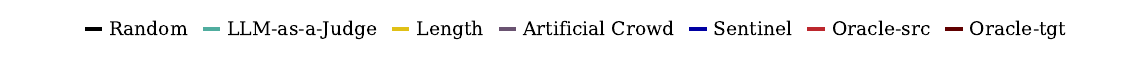}
    \includegraphics[width=\linewidth]{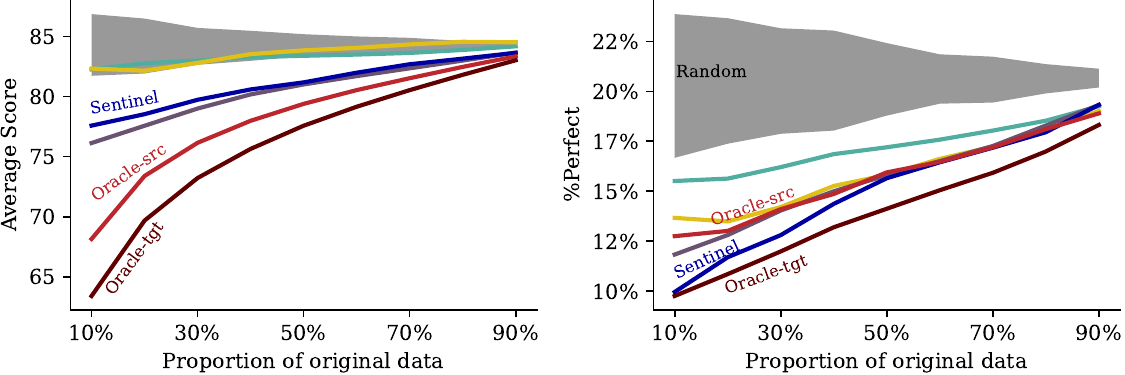}
    
    \caption{Average score and proportion of ``perfect'' source texts when creating a difficult test set. Lower values are better. Averaged across all language pairs from WMT24 on which the subset selection is simulated.
    Random selection shows 99\% confidence t-test interval from 10 runs.}
    
    \label{fig:04-variable_budget}
\end{figure*}

\section{Creating Difficult Benchmarks}
\label{sec:difficult_subset}

In this section, we use the top-performing difficulty estimators to create difficult machine translation benchmarks.
First, we show that the test set employed at the WMT 2024 General MT shared task \cite{kocmi-etal-2023-findings} is too easy for current MT models.
Then, we define the task of selecting a difficult subset of samples from a given dataset, and evaluate our estimators.

\paragraph{WMT24 is easy.}
In \Cref{fig:07-perlang_wmt24}, we report the distribution of scores assigned by human annotators to the translations of the WMT 2024 General MT shared task \citep{kocmi-etal-2024-findings}.
Notably, the best model for each language attains almost always 90 to 100 ESA points, and so does at least one system for each input text.
This is particularly concerning for the English-to-Spanish translation direction, where top systems barely made any errors.
These findings highlight the need to create more difficult benchmarks. 
To do this, we use difficulty estimators to sample difficult texts out of a larger collection.

\subsection{Setup}

Given a large set of source texts $\texts$, we aim to extract a subset $\texts' \subseteq \texts$ of maximum difficulty of size $|\texts'| = B$.
In practice, we select $B$ texts with the highest difficulty as determined by the top-performing difficulty estimators.
As discussed in \Cref{sec:results}, we exclude True Crowd estimators.

We again rely on the WMT 24 test set, which we use as $\texts$ (see \Cref{sec:experiments} for further details).
Specifically, we focus on its English source texts, which were translated into Czech, Spanish, Hindi, Icelandic, Japanese, Russian, Ukrainian, and Chinese, as well as its Czech sources translated into Ukrainian. Accordingly, since we select $B$ \textit{source} texts from $\texts$, any subset $\texts' \subseteq \texts$ necessarily contains only English and Czech source texts, while target languages are considered solely for evaluation purposes.

\paragraph{Task definition.}
We assign a single difficulty score $\hat{d}(x)$ to each sample $x \in \texts$.
For source text-only difficulty estimators, such as heuristics and learned methods, this is straightforward, as they rely solely on the given source text.
Instead, for Artificial Crowd methods, we assign to each text $x$ the average quality score of its translations, estimated using XCOMET, averaging across both the employed MT models and the target languages.
Finally, we construct $\texts'$ by selecting the $B$ most difficult source texts.

\paragraph{Evaluation.}

One goal of constructing a difficult benchmark is to identify samples where contemporary models still struggle, in order to expose their shortcomings and guide improvements in future iterations.
Therefore, we evaluate the usefulness of difficulty estimators based on the drop in the average human score obtained by the models' translations on the test set.
As additional information, we also report the proportion of ``perfect'' outputs (i.e., those that received a full score of 100/100 ESA points from human annotators) that remain in the subsampled test set.
The exact formulas for these measures are provided in \Cref{apx:implementation_details}.

\subsection{Results}
We extract several $\texts' \in \texts$ by varying the size of the subsample, and report the curves of the Average Score and \%Perfect measures in \Cref{fig:04-variable_budget}.
First, we wish to highlight that the oracles serve as a performance upper-bound only in terms of Average Score, and not in terms of \%Perfect, because they are designed to select the sources with the lowest average score, rather than the lowest \%Perfect. In this respect, oracle-src selects the sources with the lowest average difficulty score across models and target languages; instead, oracle-tgt selects a different source text for each target language, averaging difficulty scores only across MT models. 
As we can see from \Cref{fig:04-variable_budget}, text length heuristics and LLM-as-a-Judge-based methods show very close performance to random subset selection, especially in terms of Average Score. Instead, \sentineltf and Artificial Crowd perform closer to the oracles, even surpassing oracle-src in terms of \%Perfect.
In \Cref{apx:difficult_benchmark_results}, we report detailed quantitative results for the scenario where we subsample 25\% of the test set, including per-domain performance breakdowns.

Additionally, returning to the original lament of existing test sets being too easy (\Cref{fig:07-perlang_wmt24}), in Appendix \Cref{fig:07-perlang_wmt24_ext} we show how the score distribution changes when selecting difficult texts.

\begin{table}[t]
    \small
    \setlength{\tabcolsep}{1.4pt}
    \renewcommand{\arraystretch}{1.15}
    \input{generated/03-post_effect_corr}

    \caption{
    Pearson correlations between difficulty estimators and variables of interest (source length, number of errors per source word, output diversity, and proportion of unique outputs).
    All estimators assign lower values to more difficult source texts. Therefore, negative correlation indicates a positive correlation between difficulty and the variable of interest.
    See Appendix \Cref{fig:03-post_effect} for detailed visualization and \Cref{apx:implementation_details} for implementation details.
    }
    \label{tab:03-post_effect_corr}
\end{table}

\subsection{Potential Pitfalls of Selecting by Difficulty} \label{sec:pitfalls}

Selecting samples by anything other than random sampling may harbor unexpected dangers.
For example, the texts selected by a difficulty estimator might be grammatically incorrect or poorly formed. 
Here, we investigate potential pitfalls one might encounter when using difficulty as a subsampling criterion.
Specifically, we focus on:
\begin{itemize}
\item \textbf{Source length}: Longer texts are more difficult to translate compared to shorter ones. We are interested in quantifying the extent to which difficulty estimators rely on text length.
\item \textbf{Source errors}: Translating incomprehensible source texts is naturally difficult. Nevertheless, it might be undesirable to create test sets containing many garbled sources.
\item \textbf{Output diversity}: When creating a benchmark, source texts that lead to more diverse outputs are more desirable, as they help distinguish between models.
See \Cref{apx:implementation_details} for implementation details.
\end{itemize}

We present the correlation between difficulty estimators' predictions and these variables of interest in \Cref{tab:03-post_effect_corr} and Appendix \Cref{fig:03-post_effect}.
We wish to remind the reader that we defined difficulty using translation quality, meaning that lower estimators' scores indicate higher difficulty. Therefore, a negative correlation with difficulty scores should be interpreted as a strong correlation with the concept of difficulty, as estimated by our models.

As expected, all estimators show a strong negative correlation with source length, indicating they are all biased toward selecting longer outputs. 
On the contrary, this does not seem to be the case for Source errors, suggesting that our difficulty estimators do not prioritize texts containing many errors.
Finally, our results suggest that \sentineltf and Artificial Crowd select source texts that lead to more diverse outputs.

\begin{figure}[t]
    \centering
    
    \includegraphics[width=\columnwidth]{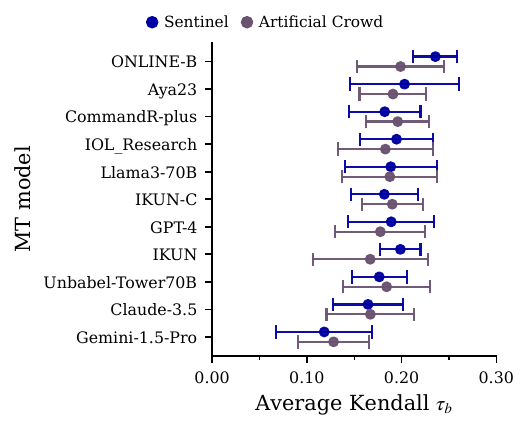}
    
    \caption{Average Kendall $\tau_b$ between difficulty estimators and the human judgments assigned to the MT models in the WMT24 test set. For each MT model and difficulty estimator, circles mark the average correlation across en$\to$X language pairs and bars report $\pm$1 standard deviation.}
    
    \label{fig:diff-alignment}
\end{figure}

\paragraph{Bias toward MT models.}
\label{sec:bias-models}

A practical concern when creating a test set using difficulty as the selection criterion is the potential bias toward certain MT models. An estimator's difficulty scores may select the texts that some MT models find more challenging than others, thereby biasing the resulting benchmark. To investigate this, we measure the alignment between our two best-performing difficulty estimators -- i.e., \sentineltf and Artificial Crowd (XCOMET) -- and the concept of translation difficulty, as defined it in \Cref{sec:difficulty}, for each MT model in isolation. Specifically, for each model and each en$\to$X translation direction, we measure the Kendall's $\tau_b$ correlation between the difficulty scores from the estimators and the human quality judgments assigned to that model's translations. Then, we average these correlations across language pairs, reporting the mean and standard deviation to capture the variability across translation directions. To ensure the averages are robust, this analysis only includes MT models that translated the WMT24 source texts into at least five target languages. The results are presented in \Cref{fig:diff-alignment}.

Overall, the mean Kendall $\tau_b$ correlations fall within a narrow range and have comparable standard deviations. This suggests that the notion of difficulty captured by both estimators aligns relatively uniformly with MT models' perceived difficulty. However, we note two exceptions: ONLINE-B exhibits a higher correlation than average with \sentineltf, whereas Gemini-1.5-Pro shows a lower average correlation with both estimators. While this is a preliminary investigation, this finding suggests that a benchmark created using these estimators could be disproportionately difficult for ONLINE-B and easier for Gemini-1.5-Pro. We leave for future work a deeper investigation into whether this discrepancy stems from a bias in the difficulty estimations or simply from variance, especially given that the correlations were averaged over a limited number of translation directions. 

\begin{table}[t]
\small
\begin{tabular}{p{7.2cm}}
\toprule
\textbf{1 (difficult):} City get a nice easy draw at home. \\\\[-0.1em]
\textbf{2 (difficult):} Alex Bregman Predicted To Betray Astros, Sign With Shocking Blue Jays \\\\[-0.1em]
\textbf{3 (difficult):} Some folks really do deserve a badge of honour for their pedantry (C8). Veronica Coyne of Springfield claims that "when bemoaning the loss of the express lane at Woolies "12 items or less," a friend told me she'd never used it on principle as it should have been "12 items or fewer.""\\\\[-0.1em]
\textbf{4 (easy):} Washington \\\\[-0.1em]
\textbf{5 (easy):} Developing the next generation of hybrid vehicles in Europe \\\\[-0.1em]
\textbf{6 (easy):} We cannot allow this to happen. This legislation is enormously unpopular. It is exactly what the American people do not want. It must not be passed by Congress. \\
\bottomrule
\end{tabular}

\captionof{example}{The most difficult and easiest English source texts from the WMT24 dataset, as selected by \sentinel.}
\label{ex:examples_sentinel}
\end{table}

\subsection{Qualitative Analysis}

We manually inspected 200 source texts, half of which were deemed easy and half difficult by \sentinelwmt, and we separated them into 10 length-based buckets.
In general, difficulty levels assigned by \sentinelwmt align well with human perception of difficulty.
Indeed, we find that difficult segments often contain complex constructions (\Cref{ex:examples_sentinel}.1), consist of incomplete sentences, such as headlines (\Cref{ex:examples_sentinel}.2), or include indirect speech (\Cref{ex:examples_sentinel}.3).
On the other hand, the segments classified as easy by \sentinelwmt are typically single words, have simple sentence structures, or are concatenations of short, simple sentences (\Cref{ex:examples_sentinel}.4 to \Cref{ex:examples_sentinel}.6).

\section{Conclusion}

In this work, we formally define the task of translation difficulty estimation and introduce the Difficulty Estimation Correlation (DEC), a dedicated measure to evaluate the performance of difficulty estimators. We conduct a comprehensive evaluation of both existing and newly proposed estimators, finding that models explicitly trained for the task significantly outperform traditional, heuristic-based methods and LLM-as-a-judge approaches.

Our analysis identifies \sentineltf as the current state-of-the-art in translation difficulty estimation.
We further validate the performance of difficulty estimators in the downstream task of creating difficult benchmarks, demonstrating that they successfully identify samples where modern MT models underperform. Also in this downstream task, \sentineltf remains the top-performing method.
Building on these findings, we develop \sentinelwmt by incorporating additional data into the training pipeline of \sentineltf, and release both models publicly.
Finally, we conduct a qualitative analysis of \sentinelwmt's predictions, offering intuitive insights into the types of texts it deems difficult.

\section*{Limitations}

\paragraph{The concept of translation difficulty.}
This work is based on the assumption that we can proxy the difficulty of a given text using the quality of the translations it produces. While we acknowledge that translation difficulty should ideally be an intrinsic property of the source -- independent of any specific translation model -- this working assumption serves our purposes, particularly for the downstream application of creating challenging machine translation benchmarks. Indeed, our research objective is to identify texts that are difficult for contemporary MT models to translate, rather than to explore the abstract, model-independent notion of translation difficulty.

\paragraph{Impact of the target language on translation difficulty.}
As discussed in Section~\ref{sec:difficulty}, the difficulty of translating a given text may depend on the target language, as corroborated theoretically by \citet{bugliarello-etal-2020-easier}.
We recognize that this aspect is only briefly mentioned, and we do not provide an investigation into this phenomenon. Nonetheless, our experiments support this hypothesis: the performance of the LLM-as-a-Judge improves when the model is given information about the target language. We therefore encourage future research to explore the influence of the target language on translation difficulty more thoroughly and to investigate how this information might be incorporated into other difficulty estimation methods effectively.

\paragraph{Using the WMT 2024 test set to analyze difficulty estimators.}
In \Cref{sec:pitfalls}, we investigated potential concerns of subsampling large data sets using our difficulty estimators. However, to do this, we used the WMT 2024 test set, which has a limitation. The sources contained in this test set were vetted by humans, making the distribution of the phenomena we investigate artificial. 
To mitigate this issue, we conduct the same analysis using a larger batch of data and report results in Appendix \Cref{fig:05-post_effect_src_diversity_wmt25}.

\section*{Ethics Statement}

We foresee no ethical issues with our work.

%% file: tables/dec_results_esa.tex
\begin{table}[t]
    \centering
    \small
    \renewcommand{\arraystretch}{1.15}
\begin{tabular}{llccr}
\toprule
 & \multicolumn{1}{l}{\textbf{Method}}
 & \hspace{-5mm}\textbf{Trans.} \hspace{-2mm}
 & \hspace{-2mm} \textbf{Lang.} \hspace{-2mm}
 & \multicolumn{1}{c}{\textbf{DEC}} \\
\midrule
\multirow[c]{3}{*}{\rotatebox[origin=c]{90}{Oracle}} & Oracle & \tick & \tick & \cellcolor{SpringGreen3!80}$1.000$ \\
& Oracle & \cross & \tick & \cellcolor{SpringGreen3!94}$0.301$ \\
& Oracle & \cross & \cross & \cellcolor{SpringGreen3!73}$0.224$ \\
\cmidrule{2-5}
\multirow[c]{3}{*}{\rotatebox[origin=c]{90}{Heuristic}} & Text Length & \cross & \cross & \cellcolor{SpringGreen3!45}\underline{$0.121$} \\
& Syntactic Complexity & \cross & \cross & \cellcolor{SpringGreen3!33}$0.080$ \\
& Word Rarity & \cross & \cross & \cellcolor{SpringGreen3!0}$-0.040$ \\
\cmidrule{2-5}
\multirow[c]{4}{*}{\rotatebox[origin=c]{90}{Learned}} & \sentineltf & \cross & \cross & \cellcolor{SpringGreen3!57}\underline{$0.182$} \\
& \sentinel & \cross & \cross & \cellcolor{SpringGreen3!54}$0.175$ \\
& PreCOMET Difficulty & \cross & \cross & \cellcolor{SpringGreen3!49}$0.153$ \\
& PreCOMET Diversity & \cross & \cross & \cellcolor{SpringGreen3!46}$0.142$ \\
\cmidrule{2-5}
\multirow[c]{4}{*}{\rotatebox[origin=c]{90}{\shortstack{LLM\\Judge}}} & Command A & \cross & \cross & \cellcolor{SpringGreen3!30}$0.072$ \\
& Command A & \cross & \tick & \cellcolor{SpringGreen3!38}\underline{$0.104$} \\
& GPT-4o & \cross & \cross & \cellcolor{SpringGreen3!32}$0.077$ \\
& GPT-4o & \cross & \tick & \cellcolor{SpringGreen3!33}$0.080$ \\
\cmidrule{2-5}
\multirow[c]{4}{*}{\rotatebox[origin=c]{90}{\shortstack{Crowd\\Based}}} & True (XCOMET) & \tick & \tick & \cellcolor{SpringGreen3!70}\underline{${0.221}$} \\
& True (MetricX) & \tick & \tick & \cellcolor{SpringGreen3!66}$0.207$ \\
& Artificial (XCOMET) & \cross & \tick & \cellcolor{SpringGreen3!54}$0.177$ \\
& Artificial (MetricX) & \cross & \tick & \cellcolor{SpringGreen3!51}$0.166$ \\
\cmidrule{2-5}
& Random & \cross & \cross & \cellcolor{SpringGreen3!11}$0.003$ \\
\bottomrule
\end{tabular}

\caption{Difficulty Estimation Correlation (DEC) achieved by each method. We categorize the methods based on the type of information they have access to.
Text-only estimators, such as the heuristic and learned ones, rely solely on the source text whose difficulty is being estimated. Instead, some methods also incorporate information of the target language (Lang.) or of the true translation included in the test set (Trans.).}
    \label{tab:dec_results_esa}
\end{table}

%% file: tables/human_mt_difficulty.tex
\begin{table}[t]
\small
\centering
\renewcommand{\arraystretch}{1.15}
\begin{tabular}{p{1.2cm}p{1.0cm}p{1.0cm}p{1.1cm}p{1.15cm}}
\toprule
 &  IOL &  GPT-4 &  Claude3.5 &  Tower70B \\
\midrule
 Human & \cellcolor{SpringGreen3!22} $0.137$ & \cellcolor{SpringGreen3!22} $0.137$ & \cellcolor{SpringGreen3!16} $0.127$ & \cellcolor{SpringGreen3!5} $0.109$ \\
 Tower70B & \cellcolor{SpringGreen3!46} $0.176$ & \cellcolor{SpringGreen3!35} $0.158$ & \cellcolor{SpringGreen3!31} $0.151$ &  \\
 Claude3.5 & \cellcolor{SpringGreen3!47} $0.178$ & \cellcolor{SpringGreen3!74} $0.221$ &  &  \\
 GPT-4 & \cellcolor{SpringGreen3!62} $0.202$ &  &  &  \\
\bottomrule
\end{tabular}
\caption{Average (across language directions) Kendall $\tau_b$ correlation matrix for human and four MT models.}
\label{tab:avg_corr_matrix_human_mt}
\end{table}

%% file: generated/03-post_effect_corr.tex
\begin{tabular}{lrrrrr} 
 \toprule

    & \multicolumn{2}{c}{\bf Source} & \multicolumn{2}{c}{\bf Diversity} & \bf Unique \\
    & \bf length & \bf errors & \bf embd & \bf chrF\hspace{0.5mm} & \bf outputs \\
    \midrule
    
Random & 
\cellcolor{red!0} $0.00$ & \cellcolor{red!0} $0.00$ & \cellcolor{red!0} $0.00$ & \cellcolor{red!0} $0.00$ & \cellcolor{red!0} $0.00$\\
LLM-as-a-Judge & 
\cellcolor{red!55} $-0.61$ & \cellcolor{red!23} $0.26$ & \cellcolor{red!16} $0.19$ & \cellcolor{red!20} $0.23$ & \cellcolor{red!54} $-0.60$\\
Length & 
\cellcolor{red!90} $-1.00$ & \cellcolor{red!22} $0.25$ & \cellcolor{red!28} $0.31$ & \cellcolor{red!22} $0.24$ & \cellcolor{red!47} $-0.52$\\
Artificial Crowd & 
\cellcolor{red!57} $-0.63$ & \cellcolor{red!3} $0.04$ & \cellcolor{red!9} $-0.11$ & \cellcolor{red!15} $-0.17$ & \cellcolor{red!41} $-0.46$\\
Sentinel & 
\cellcolor{red!59} $-0.66$ & \cellcolor{red!10} $0.12$ & \cellcolor{red!1} $-0.01$ & \cellcolor{red!7} $-0.09$ & \cellcolor{red!32} $-0.36$\\
Oracle-src & 
\cellcolor{red!19} $-0.22$ & \cellcolor{red!14} $-0.16$ & \cellcolor{red!43} $-0.47$ & \cellcolor{red!44} $-0.49$ & \cellcolor{red!25} $-0.28$\\
Oracle-tgt & 
\cellcolor{red!19} $-0.22$ & \cellcolor{red!14} $-0.16$ & \cellcolor{red!43} $-0.47$ & \cellcolor{red!44} $-0.49$ & \cellcolor{red!25} $-0.28$\\
\bottomrule 
\end{tabular}

%% file: appendix.tex
\input{tables/wmt24_stats_mqm}
\input{tables/wmt24_stats_esa}

\section[Kendall $\tau_b$]{Kendall $\bm{\tau}_{\bm{b}}$}
\label{sec:kendall}

Kendall's $\tau$ variant $b$ is defined as:
\begin{equation}
    \tau_b \;=\; \frac{C - D}{\sqrt{(C+D+T_g)(C+D+T_h)}}\,.
\end{equation}

\noindent
Here $C$ and $D$ are the numbers of \emph{concordant} and \emph{discordant} pairs when comparing gold scores $r_i$ with hypothesis scores $\hat r_i$: a pair $(i,j)$ is concordant if $(r_i-r_j)(\hat r_i-\hat r_j)>0$ and discordant if $(r_i-r_j)(\hat r_i-\hat r_j)<0$. $T_h$ counts pairs tied \emph{only} in the hypothesis ($\hat r_i=\hat r_j$ and $r_i\ne r_j$), and $T_g$ counts pairs tied \emph{only} in the gold ($r_i=r_j$ and $\hat r_i\ne\hat r_j$); pairs tied in both rankings are ignored.

We use Kendall’s $\tau_b$ rather than Pearson correlation because $\tau_b$ evaluates \emph{relative order} (ranks) and is therefore more robust to scale differences and outliers. This is important in our setting: as visible in \Cref{fig:07-perlang_wmt24}, most WMT24 segment-level scores lie in a narrow band (e.g., $[90,100]$), which magnifies the effect of outliers on Pearson, whereas $\tau_b$ depends only on pairwise rankings. This behavior is also discussed by \citet{mathur-etal-2020-tangled}.

\section{Training \sentineltf and \sentinelwmt}
\label{apx:sentinel-training}

Our new learned difficulty estimation models, \sentineltf and \sentinelwmt, follow the same architecture and training pipeline used for the \sentinel model introduced by \citet{perrella-etal-2024-guardians}. Both models are based on XLM-RoBERTa large as the backbone encoder, followed by a multi-layer feedforward network on top of the \texttt{[CLS]} token. They are trained to minimize the Mean Squared Error (MSE) between predicted and human scalar scores.

We adopt the same two-stage training approach as the \sentinel model. In the first stage, the model is trained on Direct Assessment (DA, \citealp{graham-etal-2013-continuous}) data. In the second stage, it is fine-tuned on MQM annotations. The key differences between our models and \sentinel lie in the training data used at each stage.

\begin{itemize}
    \item \textbf{Stage 1: DA training.} For \sentineltf, we extend the DA training data used by \citet{perrella-etal-2024-guardians} by including annotations from WMT 21 \cite{wenzek-etal-2021-findings}, as well as DA+SQM annotations from WMT 22 \cite{kocmi-etal-2022-findings} and WMT 23 \cite{kocmi-etal-2023-findings}. \sentinelwmt further includes the ESA annotations from WMT 24. For both model versions, we also incorporate MLQE-PE data \cite{fomicheva-etal-2022-mlqe} in the training set for this stage.
    
    \item \textbf{Stage 2: MQM fine-tuning.} In this phase, we expand the MQM training set by adding MQM annotations from WMT 23 \cite{freitag-etal-2023-results}. Unlike the \sentinel training pipeline, we do not average multiple scores per translation. Instead, we include all available annotations as individual training instances, preserving variability across raters. This applies to WMT 20 and WMT 22 MQM datasets, which include three human scores per translation \cite{freitag-etal-2021-experts, riley-etal-2024-finding}. Similarly to the first training stage on DA, in the case of the \sentinelwmt model, we also include MQM annotations from WMT 24.
\end{itemize}

\noindent
Following the approach of \citet{perrella-etal-2024-guardians}, we treat each pair consisting of a source text segment and its associated human score as an independent training instance. Since human scores are assigned to individual translations, multiple annotations may exist for the same source text. We do not combine these scores in any way but include them all in the training data for both DA and MQM stages. Training hyperparameters match those used by \citet{perrella-etal-2024-guardians} for \sentinel. All models were trained using a single NVIDIA GeForce RTX 4090 GPU. The estimated training time is approximately three GPU hours for the first (DA) stage and one GPU hour for the second (MQM) fine-tuning stage. These estimates apply to both \sentineltf and \sentinelwmt.

\section{Implementation Details}
\label{apx:implementation_details}

\begin{itemize}
\item For the word rarity heuristic, we compute word frequencies using the \href{https://github.com/rspeer/wordfreq}{\texttt{wordfreq}} Python library \cite{robyn_speer_2022_7199437}.
\item For the syntactic complexity heuristic and text length, we obtain dependency trees and corresponding tokens using \href{https://github.com/explosion/spaCy}{\texttt{spaCy}}. Specifically, we use language-specific pipelines: i) \texttt{en\_core\_web\_sm} for English, ii) \texttt{ja\_core\_news\_sm} for Japanese, and iii) \texttt{spacy\_udpipe} for Czech.
\item For the output diversity assessment (Section~\ref{sec:pitfalls}), we compute multilingual sentence embeddings from the source texts using \href{https://huggingface.co/sentence-transformers/paraphrase-multilingual-MiniLM-L12-v2}{sentence-transformers/paraphrase-multilingual-MiniLM-L12-v2} \citep{reimers-gurevych-2019-sentence}. Specifically, for each pair of translations, we measure the inner product between their multilingual sentence embeddings and their chrF score computed one against the other. 

\end{itemize}

\noindent
For the artificial crowd, we use the following:
\begin{itemize}
    \item \textbf{NLLB-moe-54B}: sparsely-gated mixture-of-experts encoder-decoder translation model \cite{nllbteam2022languageleftbehindscaling}.
    \item \textbf{Gemma-3-27B-IT}: multimodal instruction-tuned LLM from the Gemma family \cite{gemmateam2025gemma3technicalreport}.
    \item \textbf{Qwen2.5-72B-IT}: largest instruction-tuned LLM from the Qwen2.5 family \cite{qwen2025qwen25technicalreport}.
    \item \textbf{CommandA}: $111$B-parameter LLM for real-world enterprise use cases \cite{cohere2025commandaenterprisereadylarge}.
\end{itemize}

\noindent
For evaluation of \Cref{sec:difficult_subset} we use average model score and \%Perfect.
For {average model score}, given $\texts' \subseteq \texts$ and a set of models $\systems$, we report for each subset $\texts' \subseteq \texts$ with $|\texts'| = B$:
\begin{align}
\mathrm{AvgScore} = 
\frac{1}{B{\cdot}|\systems|} \sum_{\substack{x \in \texts'\\m\in \systems}} d_{m,l}(x) \,,
\label{eq:avgscore}
\end{align}
which is the average human score on the subset.
For {proportion of perfect translations}, we use:
\begin{align}
& \hspace{-3mm} \mathrm{\%Perfect} 
{=}\frac{1}{B{\cdot}|\systems|} \sum_{\substack{x \in \texts'\\m\in \systems}} \mathbbm{1}[d_{m,l}(x) {=} 100\%]
\end{align}

\begin{figure*}[t]
    \centering
    \includegraphics[width=1\linewidth]{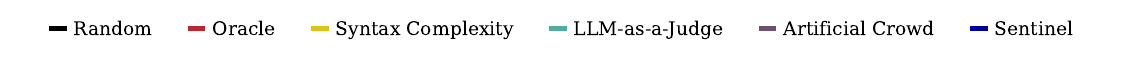}
    \includegraphics[width=1\linewidth]{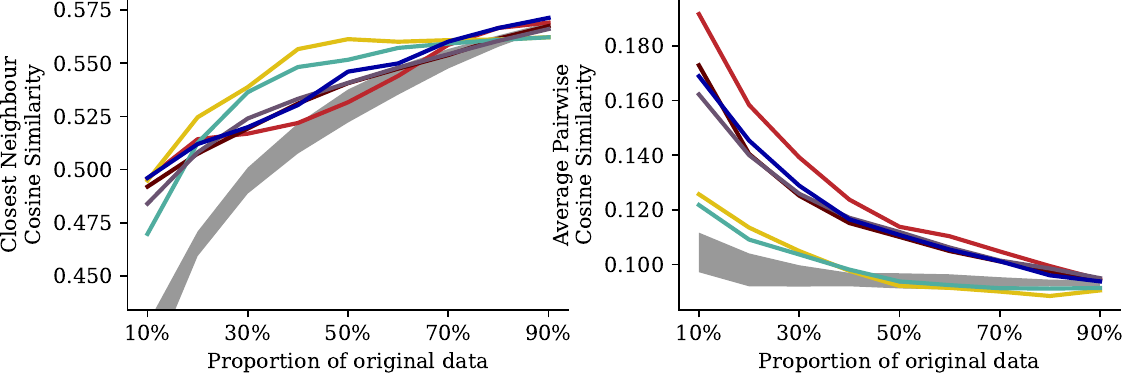}
    
    \caption{
    Average similarity between two closest (left) and any two (right) source texts in $\texts$ based on embeddings and cosine similarity. The left curves go up because the vector space saturates and nearest neighbours become closer.
    Random selection shows 99\% confidence t-test interval from 10 runs.
    }
    \label{fig:05-post_effect_src_diversity}

    \bigskip
    
    \centering
    \includegraphics[width=1\linewidth]{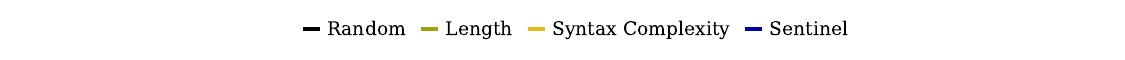}
    \includegraphics[width=1\linewidth]{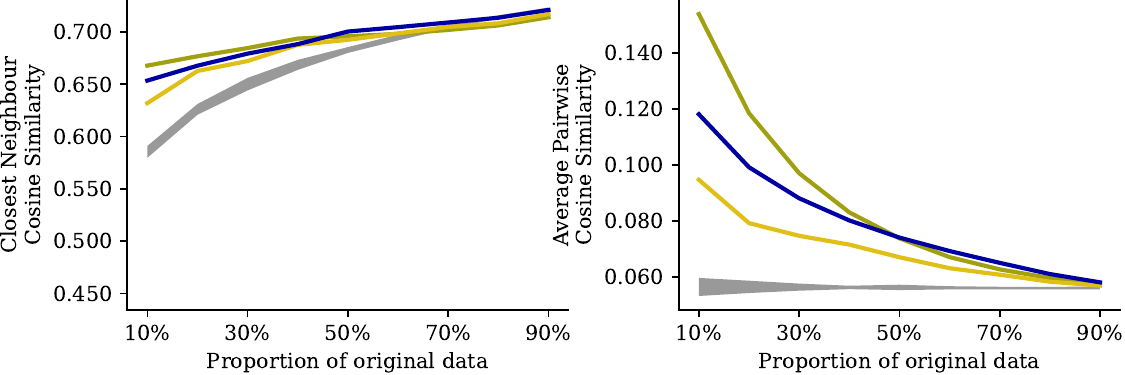}
    
    \caption{
    In contrast to \Cref{fig:05-post_effect_src_diversity}, we have collected contemporary news articles (40k segments through crawling, \citealp{banon-etal-2020-paracrawl}) to evaluate how our difficulty sampling would perform in a real world.
    Average similarity between two closest (left) and any two (right) source texts based on embeddings and cosine similarity in $\texts$ on raw 40k English segments (not WMT24). The left curves go up because the vector space saturates and nearest neighbours become closer.
    Random selection shows 99\% confidence t-test interval from 10 runs.
    LLM-as-a-Judge and Artificial crowd were not included due to compute costs.
    Oracle is not present due to the absence of model outputs and human scores.
    }
    \label{fig:05-post_effect_src_diversity_wmt25}
\end{figure*}

\begin{figure*}[t]
\centering
    \includegraphics[width=0.45\linewidth]{generated/07-perlang_wmt24_legend.pdf}
    
    \begin{minipage}{0.45\linewidth}
    \centering
    \large \bf All data\\
    \includegraphics[width=\linewidth,trim={0 4.5mm 0 0},clip]{generated/07-perlang_wmt24.pdf}
    \end{minipage}
    \begin{minipage}{0.45\linewidth}
    \centering
    \large \bf Artificial crowd (top 25\%)\\
    \includegraphics[width=\linewidth,trim={0 4.5mm 0 0},clip]{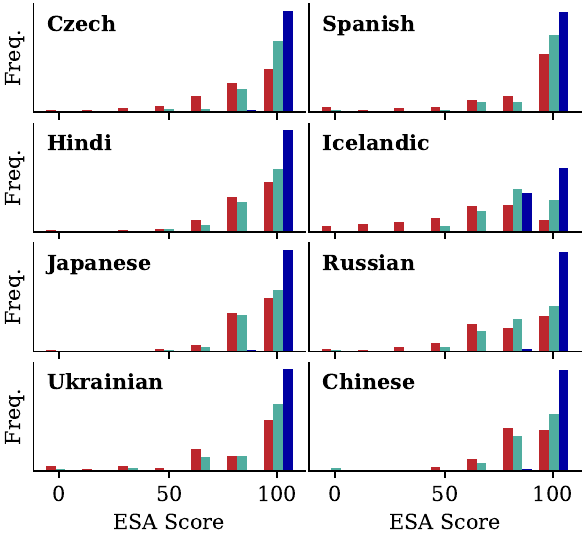}
    \end{minipage}

    \vspace{4mm}

    \begin{minipage}{0.45\linewidth}
    \centering
    \large \bf Oracle-src (top 25\%)\\
    \includegraphics[width=\linewidth,trim={0 4.5mm 0 0},clip]{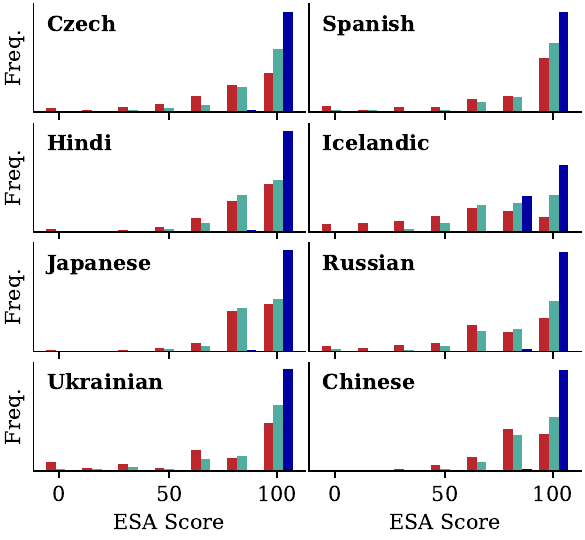}
    \end{minipage}
    \begin{minipage}{0.45\linewidth}
    \centering
    \large \bf Oracle-tgt (top 25\%)\\
    \includegraphics[width=\linewidth,trim={0 4.5mm 0 0},clip]{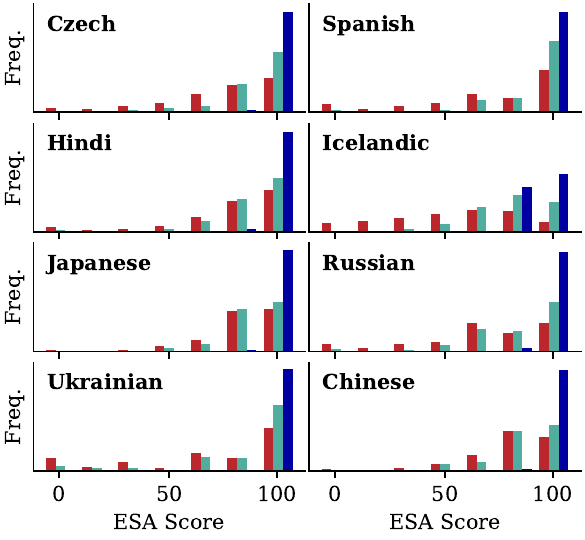}
    \end{minipage}
    
    \vspace{4mm}
    
    \begin{minipage}{0.45\linewidth}
    \centering
    \large \bf \sentineltf (top 25\%)\\
    \includegraphics[width=\linewidth]{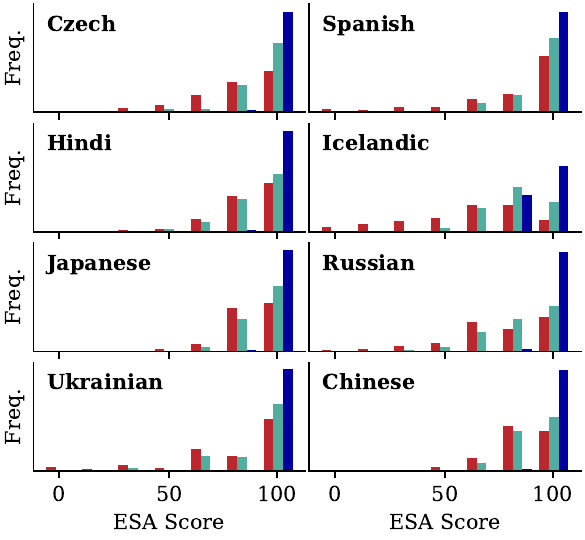}
    \end{minipage}
    \begin{minipage}{0.45\linewidth}
    \centering
    \large \bf LLM-as-a-Judge (top 25\%)\\
    \includegraphics[width=\linewidth]{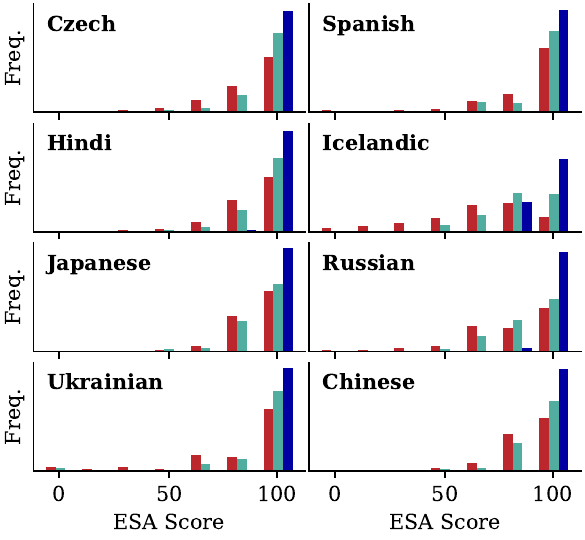}
    \end{minipage}

    \caption{Distribution of human scores for machine translation models in WMT 2024 \citep{kocmi-etal-2024-findings} of all models, top model in each language, and top model for each input segment. Subset selection methods select top 25\% most difficult segments. Extends \Cref{fig:07-perlang_wmt24}.}
    \label{fig:07-perlang_wmt24_ext}
\end{figure*}

\input{tables/dec_results_mqm}

\section{Complete Results}
\label{apx:complete_results}
\Cref{tab:dec_results_mqm} presents the difficulty estimation correlation scores of all considered methods when the ground truth is based on MQM annotations, rather than ESA.

Instead, Tables~\ref{tab:dec_results_esa_full} and \ref{tab:dec_results_mqm_full} present the per-language breakdown of all methods' difficulty estimation correlation scores on the ESA-annotated and MQM-annotated WMT24 test data, respectively. These tables also include ranks derived from statistical significance analysis. Specifically, we used the PERM-BOTH hypothesis test, introduced by \citet{deutsch-etal-2021-statistical}.  

\input{tables/dec_results_esa_full}
\input{tables/dec_results_mqm_full}

\input{tables/subset_selection_esa}
\input{tables/subset_selection_mqm}

\input{tables/subset_selection_esa_domains}
\input{tables/subset_selection_mqm_domains}

\section{Oracles} \label{apx:oracles}

Oracle methods use the true human judgments used to derive difficulty scores, as detailed in \Cref{sec:difficulty}. We consider three oracles that differ in the type of information they have access to:
\begin{itemize}
\item \textbf{Oracle (source text + target language + target translation)} assigns to each source text $x$ the true $d_{m,l}(x)$, for each $m$ and $l$.
\item \textbf{Oracle (source text + target language)} estimates the difficulty of $x$ by averaging the true $d_{m,l}(x)$ across all models ($\forall m \in \systems$), meaning that its estimates do not vary across models, but only across target languages.
\item \textbf{Oracle (source text only)} averages the true $d_{m,l}(x)$ across both models and target languages, assigning the same score to each source text regardless of target language or translator.
\end{itemize}

\begin{figure*}
    \centering
    \includegraphics[width=1\linewidth]{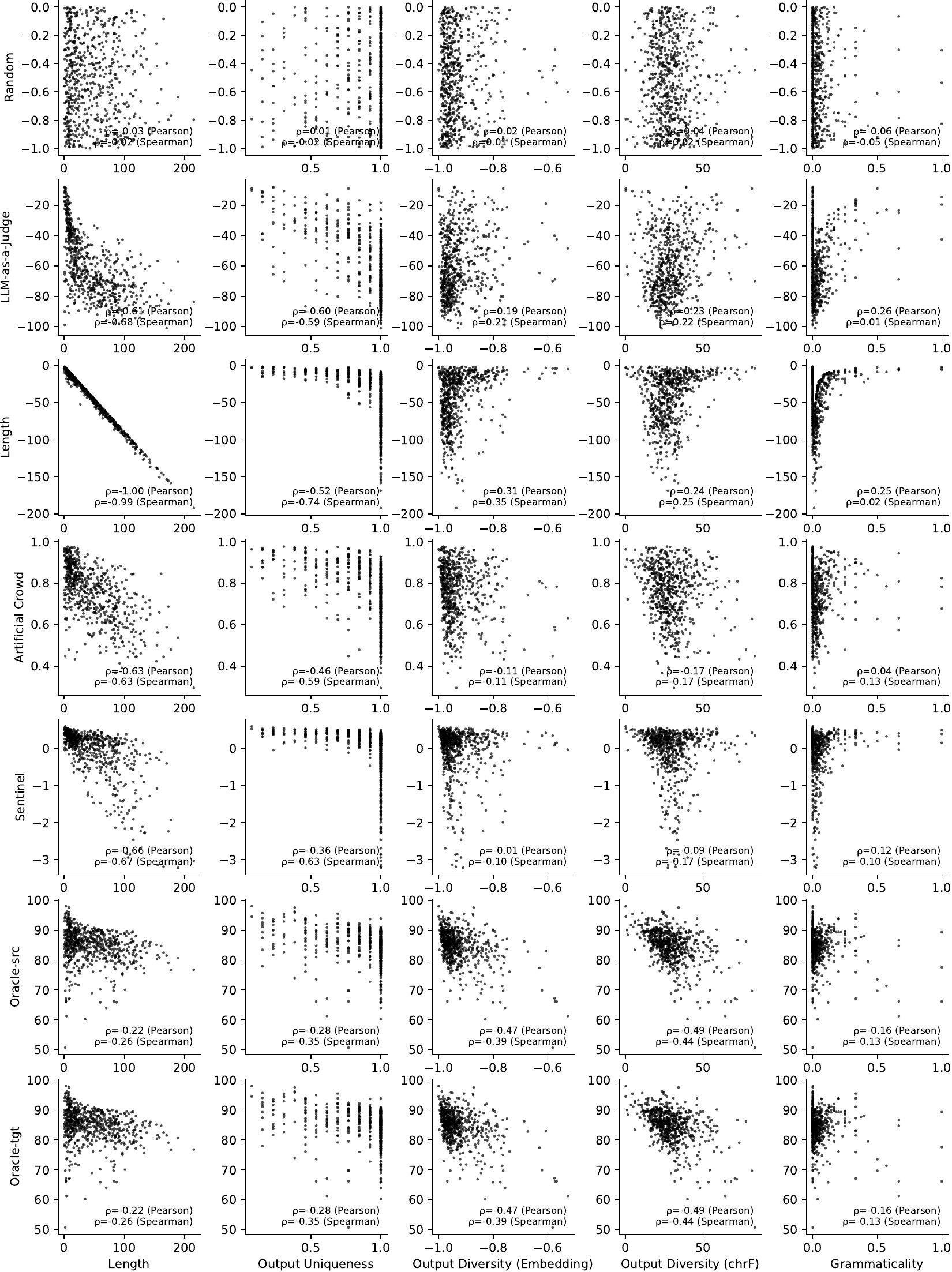}
    \caption{
        Relationship between selectors and variables of interest (source length, source number of errors per word, output diversity measured by pairwise embeddings inner product and chrF, and proportion of unique outputs).
        For all methods, lower values indicate more difficult source texts, so negative correlation implies stronger positive connection between difficulty and the target variable.
        See \Cref{tab:03-post_effect_corr} for an aggregated perspective.
    }
    \label{fig:03-post_effect}
\end{figure*}

\begin{figure*}[htbp]
\small
\centering
\begin{tabular}{p{15.5cm}}
\toprule
\bf Prompt for LLM-as-a-judge (source text only): \\[1em]

You are given a source text. Your goal is to determine the approximate proficiency level required to translate this text, based on a detailed analysis of its complexity. The final result should be reported as a single numeric score on a scale of 0 to 120, where higher numbers correspond to a higher difficulty (i.e., more advanced language proficiency requirements). You should also relate this numeric score to commonly recognized proficiency levels (e.g., A1, A2, B1, B2, C1, C2). Here is the expected mapping: 0-20 for A1 (Beginner); 21-40 for A2 (Elementary); 41-60 for B1 (Intermediate); 61-80 for B2 (Upper Intermediate); 81-100 for C1 (Advanced); 101-120 for C2 (Mastery).\\\\[-0.5em]

Instructions: First, examine the text to identify features that affect reading difficulty, including complexity of vocabulary, grammar, semantic density, and any specialized knowledge required. Then, provide a brief explanation of your reasoning for each major factor. Consider whether the text includes domain-specific terminology, cultural references, idiomatic expressions, or advanced grammatical constructions. Finally, assign a numeric score from 0 to 120 and map that score to one of the CEFR levels. Conclude with a final statement that clearly states your numeric score and the corresponding proficiency level surrounded by triple square brackets, for example [[[86, C1 (Advanced)]]]\\\\[-0.5em]

Analyze following text:\\
\{src\} \\
\end{tabular}

\vspace{10mm}
\small
\begin{tabular}{p{15.5cm}}
\bf Prompt for LLM-as-a-judge (source text + target language): \\[1em]

You are given a source text. Your goal is to determine the approximate proficiency level required to translate this text into \{target\_language\}, based on a detailed analysis of its complexity. The final result should be reported as a single numeric score on a scale of 0 to 120, where higher numbers correspond to a higher difficulty (i.e., more advanced language proficiency requirements). You should also relate this numeric score to commonly recognized proficiency levels (e.g., A1, A2, B1, B2, C1, C2). Here is the expected mapping: 0-20 for A1 (Beginner); 21-40 for A2 (Elementary); 41-60 for B1 (Intermediate); 61-80 for B2 (Upper Intermediate); 81-100 for C1 (Advanced); 101-120 for C2 (Mastery).\\\\[-0.5em]

Instructions: First, examine the text to identify features affecting the translation into \{target\_language\}, which affect reading difficulty, including complexity of vocabulary, grammar, semantic density, and any specialized knowledge required. Then, provide a brief explanation of your reasoning for each major factor. Consider whether the text includes domain-specific terminology, cultural references, idiomatic expressions, or advanced grammatical constructions. Finally, assign a numeric score from 0 to 120 and map that score to one of the CEFR levels. Conclude with a final statement that clearly states your numeric score and the corresponding proficiency level surrounded by triple square brackets, for example [[[86, C1 (Advanced)]]].\\\\[-0.5em]

Analyze following text:\\
\{src\} \\
\bottomrule
\end{tabular}

\captionof{example}{Prompts used to estimate the difficulty of a given text using LLM-as-a-judge (\Cref{sec:llm-as-a-judge}).}
\label{fig:llm_prompts}
\end{figure*}

\section{Creating Difficult Benchmarks – Quantitative Results}
\label{apx:difficult_benchmark_results}

To quantitatively evaluate the effectiveness of our difficulty estimators for constructing challenging benchmarks, we simulate a 25\% budget scenario. That is, for each method, we select the 25\% most difficult source texts from the WMT 24 test sets and assess the resulting subset using human annotations of translation quality.

Table~\ref{tab:results_main} and Table~\ref{tab:results_main_mqm} report the results of this evaluation for the ESA and MQM human annotation protocols, respectively, averaged across all language directions in the corresponding test sets. We consider two quantitative indicators: (1) {AvgScore}, the average human score assigned to the selected subset (lower indicates higher difficulty), and (2) {\%Perfect}, the proportion of model outputs in the selected subset that receive a perfect human score (lower is also better).

Results confirm the strong performance of our dedicated difficulty estimation model, \sentineltf, which achieves substantially lower AvgScore and \%Perfect values than random selection. It also achieves the best results among all automatic methods that rely solely on the source text. In particular, in Table~\ref{tab:results_main}, it is outperformed in AvgScore only by Artificial Crowd (XCOMET), a more computationally intensive approach that requires translating each source text with multiple large models and evaluating those translations using an XXL MT metric. Furthermore, Artificial Crowd methods can produce difficulty scores conditioned on the target language, unlike \sentineltf, which relies exclusively on the source text. On the other hand, \sentineltf obtains the best \%Perfect score in Table~\ref{tab:results_main}. In Table~\ref{tab:results_main_mqm}, \sentineltf outperforms all automatic methods in both AvgScore and \%Perfect, including Artificial Crowd.

As for the other automatic methods, the Text Length heuristic consistently outperforms LLM-as-a-Judge (based on Command A), despite the latter requiring significantly more computational resources. Notably, in both Table~\ref{tab:results_main} and Table~\ref{tab:results_main_mqm}, Command A only marginally improves over random selection, reinforcing the limitations of LLM-as-a-Judge methods already observed in Table~\ref{tab:dec_results_esa}.

Tables~\ref{tab:results_fine_grained} and~\ref{tab:results_fine_grained_mqm} provide a fine-grained breakdown of results across the WMT 24 domains (News, Social, Literary, and Speech) for the ESA and MQM test sets, respectively. These results show that the overall patterns hold consistently across domains. While absolute performance varies, \sentineltf and Artificial Crowd achieve the strongest results in nearly all domain-specific evaluations.

This analysis supports the practical utility of difficulty estimation for controlled test set construction and confirms that learned estimators such as \sentineltf offer effective and reliable means for identifying source segments where MT systems are more likely to struggle.

\section{Related work for Benchmark Creation}

We extend the related work in \Cref{sec:related_work} by discussion on previous attempts to automatically create challenging subsets.

\citet[tinyBenchmarks]{pmlr-v235-maia-polo24a} and \citet{rodriguez-etal-2021-evaluation} make heavy use of Item Response Theory \citep{santor1998progress}, which is a set of statistical models for educational testing of human subjects.
However, this is not applicable to machine translation, where the quality of the output is represented as a continuous score.
Other works \citep{ni2024mixevalderivingwisdomcrowd,ni2024mixevalxanytoanyevaluationsrealworld,Ruan_Pu_Gao_Wan_Zhu_2024,zouhar2025subset}
attempt to be more broadly applicable to natural language generation tasks, though their optimization goals are usually efficient testing (i.e. obtaining the same model ranking with fewer evaluated examples) rather than creating difficult testsets.

For machine translation specifically, \citet{zhan-etal-2021-difficulty} use proxy of machine translation difficulty to inform better evaluation.
Again, \citet{zouhar-etal-2025-ai} automatically remove examples that are too easy from the evaluation set, corresponding to our True Crowd with quality estimation.

%% file: tables/wmt24_stats_mqm.tex
\begin{table}[t]
    \centering
    \small
    \begin{tabular}{lrrr}
        \toprule
        &  \multicolumn{1}{c}{\langpair{en}{de}} & \multicolumn{1}{c}{\langpair{en}{es}} &  \multicolumn{1}{c}{\langpair{ja}{zh}} \\

        \cmidrule(lr){2-4}
        
        \#Source texts & 486 & 622 & 559 \\
        \#Translators & 19 & 15 & 15 \\

        \bottomrule
    \end{tabular}
    \caption{Statistics of the test set released at the WMT 2024 Metrics Shared Task \cite{freitag-etal-2024-llms}. ``\#Source texts'' indicates the number of source texts in the test set, and ``\#Translators'' indicates the number of available translations for each source text.}
    \label{tab:wmt24_stats_mqm}
\end{table}

%% file: tables/wmt24_stats_esa.tex
\begin{table*}[t]
    \centering
    \small
    \begin{tabular}{lrrrrrrrrrr}
        \toprule
        &  \multicolumn{1}{c}{\langpair{en}{es}} & \multicolumn{1}{c}{\langpair{en}{hi}} &  \multicolumn{1}{c}{\langpair{en}{is}} & \multicolumn{1}{c}{\langpair{en}{ja}}         &  \multicolumn{1}{c}{\langpair{en}{ru}} & \multicolumn{1}{c}{\langpair{en}{uk}}         &  \multicolumn{1}{c}{\langpair{en}{zh}} & \multicolumn{1}{c}{\langpair{en}{cs}} & \langpair{cs}{uk} \\

        \cmidrule(lr){2-10}
        
        \#Source texts & 634 & 634 & 634 & 634 & 634 & 634 & 634 & 634 & 1954  \\
        \#Translators & 14 & 11 & 11 & 13 & 14 & 11 & 13 & 16 & 12 \\

        \bottomrule
    \end{tabular}
    \caption{Statistics of the test set released at the WMT 2024 General Machine Translation Shared Task \cite{kocmi-etal-2024-findings}. ``\#Source texts'' indicates the number of source texts in the test set, and ``\#Translators'' indicates the number of available translations for each source text.}
    \label{tab:wmt24_stats_esa}
\end{table*}

%% file: tables/dec_results_mqm.tex
\begin{table*}[t]
    \centering
    \small

    \begin{tabular}{llccr}
\toprule
 & \multicolumn{1}{l}{\textbf{Method}}
 & \multicolumn{1}{c}{\textbf{System}}
 & \multicolumn{1}{c}{\textbf{Lang}}
 & \multicolumn{1}{c}{\textbf{DEC}} \\
\midrule

\multirow[c]{3}{*}{\rotatebox[origin=c]{90}{Oracle}} & Oracle & \tick & \tick & $1.000$ \\
& Oracle (source text + target lang) & \cross & \tick & $0.430$ \\
& Oracle (source text only) & \cross & \cross & $0.404$ \\
\cmidrule{2-5}
\multirow[c]{3}{*}{\rotatebox[origin=c]{90}{Heuristic}} & Text Length & \cross & \cross & \underline{$0.222$} \\
& Syntactic Complexity & \cross & \cross & $0.170$ \\
& Word Rarity & \cross & \cross & $-0.052$ \\
\cmidrule{2-5}
\multirow[c]{4}{*}{\rotatebox[origin=c]{90}{Learned}} & \sentineltf & \cross & \cross & \underline{$0.246$} \\
& \sentinel & \cross & \cross & $0.235$ \\
& PreCOMET Difficulty & \cross & \cross & $0.169$ \\
& PreCOMET Diversity & \cross & \cross & $0.167$ \\
\cmidrule{2-5}
\multirow[c]{4}{*}{\rotatebox[origin=c]{90}{\shortstack{LLM\\Judge}}} & Command A (source text only) & \cross & \cross & $0.114$ \\
& Command A (source text + target lang) & \cross & \tick & \underline{$0.120$} \\
& GPT-4o (source text only) & \cross & \cross & $0.090$ \\
& GPT-4o (source text + target lang) & \cross & \tick & $0.090$ \\
\cmidrule{2-5}
\multirow[c]{4}{*}{\rotatebox[origin=c]{90}{\shortstack{Crowd\\Based}}} & True (XCOMET-QE-XXL) & \tick & \tick & \underline{$\mathbf{0.278}$} \\
& True (MetricX-24-Hybrid-QE-XXL) & \tick & \tick & $0.248$ \\
& Artificial (XCOMET-QE-XXL) & \cross & \tick & $0.207$ \\
& Artificial (MetricX-24-Hybrid-QE-XXL) & \cross & \tick & $0.185$ \\
\cmidrule{2-5}
& Random & \tick & \tick & $0.002$ \\

\bottomrule
\end{tabular}

\caption{Difficulty Estimation Correlation (DEC) achieved by each method on the MQM-annotated WMT24. We categorize the methods based on the type of information they have access to.
Text-only estimators, such as the heuristic and learned ones, rely solely on the source text whose difficulty is being estimated. Instead, some methods also incorporate information on the target language of translation, while others further leverage knowledge of the specific translator who produced the translations in the test set.}
\label{tab:dec_results_mqm}
\end{table*}

%% file: tables/dec_results_esa_full.tex
\begin{sidewaystable*}[t]
    \centering
    \small
\begin{tabular}{lrrrrrrrrrrr}
\toprule
 & \multicolumn{2}{c}{Average} & \multicolumn{1}{c}{\langpair{cs}{uk}} & \multicolumn{1}{c}{\langpair{en}{cs}} & \multicolumn{1}{c}{\langpair{en}{es}} & \multicolumn{1}{c}{\langpair{en}{hi}} & \multicolumn{1}{c}{\langpair{en}{is}} & \multicolumn{1}{c}{\langpair{en}{ja}} & \multicolumn{1}{c}{\langpair{en}{ru}} & \multicolumn{1}{c}{\langpair{en}{uk}} & \multicolumn{1}{c}{\langpair{en}{zh}} \\
 & Rank & DEC & DEC & DEC & DEC & DEC & DEC & DEC & DEC & DEC & DEC \\
\midrule
Oracle & 1 & 1.000 & 1.000 & 1.000 & 1.000 & 1.000 & 1.000 & 1.000 & 1.000 & 1.000 & 1.000 \\
Oracle (source text + target lang) & 2 & 0.301 & 0.298 & 0.303 & 0.280 & 0.271 & 0.381 & 0.252 & 0.320 & 0.305 & 0.302 \\
Oracle (source text only) & 3 & 0.224 & 0.298 & 0.246 & 0.201 & 0.176 & 0.238 & 0.172 & 0.253 & 0.213 & 0.218 \\
True Crowd (XCOMET-QE-XXL) & 3 & 0.221 & 0.203 & 0.271 & 0.194 & 0.213 & 0.273 & 0.173 & 0.238 & 0.195 & 0.233 \\
True Crowd (MetricX-24-Hybrid-QE-XXL) & 4 & 0.207 & 0.211 & 0.256 & 0.184 & 0.212 & 0.221 & 0.176 & 0.229 & 0.175 & 0.203 \\
\sentineltf & 5 & 0.182 & 0.167 & 0.216 & 0.169 & 0.173 & 0.220 & 0.142 & 0.204 & 0.150 & 0.197 \\
Artificial Crowd (XCOMET-QE-XXL) & 6 & 0.177 & 0.175 & 0.192 & 0.146 & 0.179 & 0.240 & 0.128 & 0.194 & 0.160 & 0.183 \\
\sentinel & 6 & 0.175 & 0.164 & 0.205 & 0.159 & 0.171 & 0.223 & 0.118 & 0.201 & 0.141 & 0.190 \\
Artificial Crowd (MetricX-24-Hybrid-QE-XXL) & 7 & 0.166 & 0.181 & 0.174 & 0.121 & 0.162 & 0.247 & 0.128 & 0.180 & 0.136 & 0.162 \\
PreCOMET Difficulty & 8 & 0.153 & 0.137 & 0.193 & 0.131 & 0.139 & 0.188 & 0.120 & 0.166 & 0.131 & 0.170 \\
PreCOMET Diversity & 9 & 0.142 & 0.059 & 0.167 & 0.134 & 0.129 & 0.213 & 0.120 & 0.159 & 0.130 & 0.165 \\
Text Length & 10 & 0.121 & 0.024 & 0.133 & 0.129 & 0.143 & 0.206 & 0.078 & 0.142 & 0.100 & 0.132 \\
LLM-as-a-Judge (Command A, tgt-based) & 11 & 0.104 & 0.077 & 0.100 & 0.098 & 0.120 & 0.190 & 0.068 & 0.117 & 0.072 & 0.097 \\
Syntactic Complexity & 12 & 0.080 & 0.018 & 0.078 & 0.072 & 0.112 & 0.181 & 0.035 & 0.090 & 0.050 & 0.079 \\
LLM-as-a-Judge (GPT-4o, tgt-based) & 12 & 0.080 & 0.061 & 0.067 & 0.072 & 0.116 & 0.179 & 0.035 & 0.079 & 0.037 & 0.071 \\
LLM-as-a-Judge (GPT-4o, src-based) & 13 & 0.077 & 0.038 & 0.066 & 0.072 & 0.111 & 0.188 & 0.029 & 0.083 & 0.036 & 0.071 \\
LLM-as-a-Judge (Command A, src-based) & 14 & 0.072 & 0.045 & 0.063 & 0.062 & 0.103 & 0.169 & 0.026 & 0.079 & 0.029 & 0.072 \\
Random & 15 & 0.003 & -0.001 & 0.004 & 0.004 & -0.008 & 0.010 & 0.008 & 0.005 & 0.008 & 0.000 \\
Word Rarity & 16 & -0.040 & 0.016 & -0.034 & -0.044 & -0.065 & -0.093 & -0.032 & -0.043 & -0.022 & -0.043 \\
\bottomrule
\end{tabular}
    \caption{Difficulty Estimation Correlation (DEC) achieved by each method, per language, on the ESA-annotated WMT24. Ranks represent clusters of statistical significance and are computed following \citet{freitag-etal-2024-llms}, which leverage the PERM-BOTH hypothesis test introduced by \citet{deutsch-etal-2021-statistical}.}
    \label{tab:dec_results_esa_full}
\end{sidewaystable*}

%% file: tables/dec_results_mqm_full.tex
\begin{table*}[t]
    \centering
    \small
\begin{tabular}{lrrrrr}
\toprule
 & \multicolumn{2}{c}{Average} & \multicolumn{1}{c}{\langpair{en}{de}} & \multicolumn{1}{c}{\langpair{en}{es}} & \multicolumn{1}{c}{\langpair{ja}{zh}} \\
 & Rank & DEC & DEC & DEC & DEC \\
\midrule
Oracle & 1 & 1.000 & 1.000 & 1.000 & 1.000 \\
Oracle (source text + target lang) & 2 & 0.430 & 0.505 & 0.280 & 0.503 \\
Oracle (source text only) & 3 & 0.404 & 0.488 & 0.221 & 0.503 \\
True Crowd (XCOMET-QE-XXL) & 4 & 0.278 & 0.309 & 0.208 & 0.315 \\
True Crowd (MetricX-24-Hybrid-QE-XXL) & 5 & 0.248 & 0.268 & 0.192 & 0.284 \\
\sentineltf & 5 & 0.246 & 0.278 & 0.168 & 0.291 \\
\sentinel & 6 & 0.235 & 0.273 & 0.165 & 0.268 \\
Text Length & 7 & 0.222 & 0.262 & 0.147 & 0.256 \\
Artificial Crowd (XCOMET-QE-XXL) & 8 & 0.207 & 0.243 & 0.159 & 0.220 \\
Artificial Crowd (MetricX-24-Hybrid-QE-XXL) & 9 & 0.185 & 0.209 & 0.145 & 0.201 \\
Syntactic Complexity & 10 & 0.170 & 0.158 & 0.073 & 0.278 \\
PreCOMET Difficulty & 10 & 0.169 & 0.219 & 0.129 & 0.159 \\
PreCOMET Diversity & 10 & 0.167 & 0.241 & 0.143 & 0.117 \\
LLM-as-a-Judge (Command A, tgt-based) & 11 & 0.120 & 0.122 & 0.088 & 0.150 \\
LLM-as-a-Judge (Command A, src-based) & 11 & 0.114 & 0.117 & 0.060 & 0.165 \\
LLM-as-a-Judge (GPT-4o, tgt-based) & 12 & 0.090 & 0.096 & 0.064 & 0.110 \\
LLM-as-a-Judge (GPT-4o, src-based) & 12 & 0.090 & 0.111 & 0.049 & 0.109 \\
Random & 13 & 0.002 & 0.003 & 0.004 & 0.000 \\
Word Rarity & 14 & -0.052 & -0.114 & -0.043 & 0.001 \\
\bottomrule
\end{tabular}
    \caption{Difficulty Estimation Correlation (DEC) achieved by each method, per language, on the MQM-annotated WMT24. Ranks represent clusters of statistical significance and are computed following \citet{freitag-etal-2024-llms}, which leverage the PERM-BOTH hypothesis test introduced by \citet{deutsch-etal-2021-statistical}.}
    \label{tab:dec_results_mqm_full}
\end{table*}

%% file: tables/subset_selection_esa.tex
\begin{table*}[t]
\centering
\small
\begin{tabular}{lcccc}
\toprule
\bf Method & \bf AvgScore & \bf \%Perfect \\
\midrule \null
\input{generated/01-eval_all_esa} \\[-1em]
\bottomrule
\end{tabular}
\caption{
Comparison of methods for selecting the most difficult 25\% of samples from the ESA test set, evaluated using (1) the average human score on the selected subset and (2) the proportion of model outputs in the selected subset that achieve a perfect human score. 
Results are calculated per language pair and then averaged. 
The entire test set has an average score (AvgScore) of 84.4 and a percentage of perfect outputs (\%Perfect) of 20.7\%.}
\label{tab:results_main}
\end{table*}

%% file: generated/01-eval_all_esa.tex
                        Random & 84.4 & 21.0\% \\
     Oracle (source text only) & 74.9 & 13.3\% \\
Oracle (source text + target lang) & 71.6 & 11.4\% \\
                   Text Length & 82.7 & 14.1\% \\
          \sentineltf & 79.1 & 12.1\% \\
Artificial Crowd (XCOMET-QE-XXL) & 78.3 & 13.3\% \\
Command A (source text + target lang) & 83.0 & 16.1\% \\

%% file: tables/subset_selection_mqm.tex
\begin{table*}[t]
\centering
\small
\begin{tabular}{lcccc}
\toprule
\bf Method & \bf AvgScore & \bf \%Perfect \\
\midrule \null
\input{generated/01-eval_all_mqm} \\[-1em]
\bottomrule
\end{tabular}
\caption{
Comparison of methods for selecting the most difficult 25\% of samples from the MQM test set, evaluated using (1) the average human score on the selected subset and (2) the proportion of model outputs in the selected subset that achieve a perfect human score. 
Results are calculated per language pair and then averaged. 
The entire test set has an average score (AvgScore) of -2.5 and a percentage of perfect outputs (\%Perfect) of 57.7\%.}
\label{tab:results_main_mqm}
\end{table*}

%% file: generated/01-eval_all_mqm.tex
                        Random & -2.5 & 58.8\% \\
     Oracle (source text only) & -6.6 & 32.7\% \\
Oracle (source text + target lang) & -6.8 & 30.5\% \\
                   Text Length & -4.5 & 43.6\% \\
          \sentineltf & -5.1 & 39.6\% \\
Artificial Crowd (XCOMET-QE-XXL) & -4.4 & 43.8\% \\
Command A (source text + target lang) & -3.1 & 51.1\% \\

%% file: tables/subset_selection_esa_domains.tex
\begin{table*}[t]
  \centering
  \scriptsize                %
  \setlength{\tabcolsep}{3pt} %
  \begin{adjustbox}{width=\textwidth}
    \begin{tabular}{
      l
      *{4}{S[table-format=2.1]}   %
      *{4}{S[table-format=2.1]}   %
    }
      \toprule
        & \multicolumn{4}{c}{\bf AvgScore}
        & \multicolumn{4}{c}{\bf \%Perfect} \\
      \cmidrule(lr){2-5}
      \cmidrule(lr){6-9}
      \bf Method
        & {News} & {Social} & {Literary} & {Speech}
        & {News} & {Social} & {Literary} & {Speech} \\
      \midrule
      \input{generated/01-eval_all_domains_esa.tex} \\[-1em]
      \bottomrule
    \end{tabular}
  \end{adjustbox}
  \caption{
    Fine-grained evaluation of the most difficult 25\% of test set samples from the ESA test set, selected independently for each domain (News, Social, Literary, Speech) and averaged across the language pairs.
    Results are shown for AvgScore (average human score) and \%Perfect (proportion of model outputs with a perfect human score).
}
  \label{tab:results_fine_grained}
\end{table*}

%% file: generated/01-eval_all_domains_esa.tex
Random & 86.5 & 84.7 & 84.7 & 80.3 & 19.3\% & 22.6\% & 19.7\% & 12.3\% \\
Oracle (source text only) & 82.5 & 75.8 & 76.0 & 71.1 & 14.1\% & 16.9\% & 11.0\% & 7.0\% \\
Oracle (source text + target lang) & 79.6 & 71.3 & 72.9 & 68.3 & 11.8\% & 13.7\% & 8.0\% & 4.7\% \\
Text Length & 84.6 & 83.1 & 78.4 & 82.0 & 15.0\% & 15.5\% & 9.7\% & 11.8\% \\
\sentineltf & 84.1 & 80.2 & 78.7 & 77.5 & 14.4\% & 15.1\% & 10.0\% & 8.6\% \\
Artificial Crowd (XCOMET-QE-XXL) & 84.6 & 79.6 & 77.6 & 75.6 & 15.3\% & 16.6\% & 11.9\% & 8.1\% \\
Command A (source text + target lang) & 84.9 & 82.1 & 79.7 & 78.8 & 15.5\% & 17.4\% & 10.6\% & 10.2\% \\

%% file: tables/subset_selection_mqm_domains.tex
\begin{table*}[t]
  \centering
  \scriptsize                %
  \setlength{\tabcolsep}{3pt} %
  \begin{adjustbox}{width=\textwidth}
    \begin{tabular}{
      l
      *{4}{S[table-format=2.1]}   %
      *{4}{S[table-format=2.1]}   %
    }
      \toprule
        & \multicolumn{4}{c}{\bf AvgScore}
        & \multicolumn{4}{c}{\bf \%Perfect} \\
      \cmidrule(lr){2-5}
      \cmidrule(lr){6-9}
      \bf Method
        & {News} & {Social} & {Literary} & {Speech}
        & {News} & {Social} & {Literary} & {Speech} \\
      \midrule
      \input{generated/01-eval_all_domains_mqm.tex} \\[-1em]
      \bottomrule
    \end{tabular}
  \end{adjustbox}
  \caption{
    Fine-grained evaluation of the most difficult 25\% of test set samples from the MQM test set, selected independently for each domain (News, Social, Literary, Speech) and averaged across the language pairs.
    Results are shown for AvgScore (average human score) and \%Perfect (proportion of model outputs with a perfect human score).
  }
  \label{tab:results_fine_grained_mqm}
\end{table*}

%% file: generated/01-eval_all_domains_mqm.tex
Random & -1.4 & -1.4 & -3.5 & -5.5 & 64.6\% & 68.9\% & 56.5\% & 37.5\% \\
Oracle (source text only) & -4.5 & -3.1 & -5.9 & -10.5 & 37.0\% & 45.6\% & 40.9\% & 24.6\% \\
Oracle (source text + target lang) & -4.7 & -3.4 & -5.9 & -11.0 & 33.6\% & 41.8\% & 40.7\% & 22.2\% \\
Text Length & -3.3 & -2.0 & -5.2 & -6.0 & 47.4\% & 58.5\% & 46.4\% & 37.2\% \\
\sentineltf & -2.7 & -2.2 & -4.9 & -7.1 & 48.7\% & 56.5\% & 48.5\% & 30.2\% \\
Artificial Crowd (XCOMET-QE-XXL) & -2.5 & -2.0 & -3.5 & -7.6 & 51.0\% & 57.8\% & 50.4\% & 30.4\% \\
Command A (source text + target lang) & -2.3 & -1.6 & -4.1 & -4.4 & 50.0\% & 64.2\% & 51.2\% & 40.5\% \\